\documentclass{article}

\usepackage[final, nonatbib]{neurips_robustseq_2022}

\usepackage[utf8]{inputenc} %
\usepackage[T1]{fontenc}    %
\usepackage{hyperref}       %
\usepackage{url}            %
\usepackage{booktabs}       %
\usepackage{amsfonts}       %
\usepackage{nicefrac}       %
\usepackage{microtype}      %

\usepackage{multirow}
\usepackage{subcaption}
\usepackage{graphicx}
\usepackage{xr}
\usepackage{placeins}

\usepackage{amsmath}
\usepackage{amssymb}

\hypersetup{colorlinks=true}

\newcommand{\condprob}[2]{
    \mathbb{P}(#1 \, | \, #2)
}

\title{Comparison of Uncertainty Quantification with Deep Learning in Time Series Regression}

\author{%
    \parbox{\linewidth}{
        Levente Foldesi$^1$, Matias Valdenegro-Toro$^1$\\
    }\\
    ~\\
    \parbox{\linewidth}{
        $^1$ Department of AI, University of Groningen, 9747 AG Groningen, The Netherlands.\\
        ~\\
        \texttt{l.foldesi@student.rug.nl}, \, \texttt{m.a.valdenegro.toro@rug.nl}
    }
}

\begin{document}

\maketitle

\begin{abstract}
    Increasingly high-stakes decisions are made using neural networks in order to make predictions. Specifically, meteorologists and hedge funds apply these techniques to time series data. When it comes to prediction, there are certain limitations for machine learning models (such as lack of expressiveness, vulnerability of domain shifts and overconfidence) which can be solved using uncertainty estimation. There is a set of expectations regarding how uncertainty should ``behave". For instance, a wider prediction horizon should lead to more uncertainty or the model's confidence should be proportional to its accuracy. In this paper, different uncertainty estimation methods are compared to forecast meteorological time series data and evaluate these expectations. The results show how each uncertainty estimation method performs on the forecasting task, which partially evaluates the robustness of predicted uncertainty.
\end{abstract}

\section{Introduction}
\label{sec:introduction}
Neural networks are often used to tackle complex tasks in fields such as robotics \cite{sunderhauf2018limits}, computer vision \cite{valdenegro2021find}, stock market/weather prediction \cite{schultz2021can}, etc. Specifically, time series data with high volatility, rely heavily on neural networks when it comes to predicting future values. Nowadays many application specialists use AI for different applications that need robust predictions. Time series data is very important for some applications involving streaming or time varying data, such as sequences, weather or environmental variable predictions, and price modeling. These applications also require uncertainty quantification at the output for robust decision making \cite{sangiorgio2020robustness}.

However, there are still plenty of limitations that need to be taken into account. The most prominent issues described by \cite{Jakob} are: "black boxes" which suggests the lack of expressiveness and interpretability of the algorithm, vulnerability to domain shifts and struggle with identifying in-domain and out-of-domain data. This means that, for example, if an image classification model is trained to distinguish between cars and bikes, but during the testing a plane is given, the model will predict something instead of noticing the out-of-domain data. Neural network outputs do not provide uncertainty (how sure the given output is) and often make overconfident predictions, and are prone to adversarial attacks, which could cause them to malfunction.

The aforementioned limitations can be improved by estimating output uncertainty \cite{hendrycks2019augmix}. Given these estimates, a human supervisor could decide whether to use the network output (at least parts of the results) or disregard them. In particular time series data impose additional constraints on how predictive uncertainty can be interpreted, for example, it is expected that uncertainty increases with the prediction horizon \cite{menezes2008long}. The far future is harder to predict than the near future. Since all uncertainty methods are approximations of the true posterior distribution, this behavior is not ensured. This can be seen as a robustness concept applied to the uncertain predictions specifically for time series data.

In this paper, we make a comparison of uncertainty quantification methods for time series regression using multiple metrics measuring different effects, across two datasets of environmental and weather data. The contributions of this work are the comparison of uncertainty quantification methods for time series data, including a feed forward multi-layer perceptron (MLP) and a Long-Short Term Memory recurrent network (LSTM), a qualitative evaluation methodology that includes prediction horizon quality, a ranking of the evaluated methods, calibration error, and mean squared error. Our results provide insights that go beyond standard evaluation of time series methods with uncertainty.

\section{Experimental Setup}
\textbf{Datasets}. We used two weather and environmental/meteorology data sets.

\textit{PM2.5 dataset} Prediction is made with regard to PM2.5 concentration. This is also called fine particulate matter, which has a harmful effect on human health \cite{pope}. The dataset can be found at \url{https://archive.ics.uci.edu/ml/datasets/Beijing+PM2.5+Data}. This dataset consists of 43825 time series with 8 features measured for each time frame (year, month, day and hour). The 8 features are the following: PM2.5 concentration, Dew Point, Temperature, Pressure, Combined wind direction, Cumulated wind speed, Cumulated hours of snow, and Cumulated hours of rain.

\textit{Air Pressure dataset} This is a weather time series data that can be downloaded at \url{https://www.bgc-jena.mpg.de/wetter/}. In this data set, 14 different features were included: Air pressure, Air temperature, Potential temperature, Dew point temperature, Relative humidity, Saturation water vapor pressure, Actual water vapor pressure, Water vapor pressure, Deficit specific humidity, Water vapor concentration, Air density, Wind velocity, Maximum wind velocity and Wind direction. The data points are measured in every 10 minutes from 2009-2016. Since prediction regarding meteorological data is usually done by hours and the previous data set was also measured in each hour, the data set was converted from minute representation to hour representation by taking every $6^{th}$ data point only.

Both datasets have the same pre-processing, transforming to a time series with 12 past timesteps, to predict a single future timestep. To evaluate prediction horizon, we train models with up to 12 future timesteps, to be consistent with the number of input timesteps.

\textbf{Uncertainty Quantification Methods}. We compare multiple methods, namely a baseline model with only aleatoric uncertainty using the Gaussian NLL loss (Equation \ref{eq:NLL}), MC-Dropout \cite{gal}, MC-DropConnect \cite{mobiny2021dropconnect}, Flipout \cite{Wen}, Bayes by Backpropagation (BBB) \cite{Charles}, and Ensembles \cite{Balaji}. We provide short descriptions of these methods in the appendix.

\textbf{Models}. We use two models. One is an MLP with two hidden Dense layers with 32 neurons using the ReLU activation, and a single output layer with one neuron. The second model similar but replacing all Dense layers with LSTM \cite{hochreiter}, producing a recurrent architecture. To combine models with specific uncertainty quantification methods, all Dense layers are replaced with corresponding uncertainty layers (Dense with DropConnectDense, or FlipoutDense, VariationalDense, etc), except for the LSTM layers, which are standard LSTM layers since there are no variations of LSTM with uncertainty.

\textbf{Training}. Loss function is the mean squared error. The Adam optimizer was used for training with a learning rate of 0.001. The number of training epochs was set to 100. Each model is combined with an uncertainty quantification method, which produces mean $\mu(x)$ and standard deviation $\sigma(x)$ predictions for input $x$. The $\sigma(x)$ is the uncertainty value associated with the prediction.

\textbf{Metrics}. We used multiple metrics to evaluate different aspects of prediction quality. Mean absolute percentage error (MAPE) for direct prediction error, Mean squared error (MSE) for a global error metric, $R^2$ score for overall prediction quality, expected calibration error (ECE) for quality of the uncertainty output \cite{guo2017calibration}, and negative-log likelihood (NLL, Eq \ref{eq:NLL}) for an overall combination of error and uncertainty outputs \cite{kendall2017uncertainties}. NLL is a proper scoring rule that scores both prediction and its uncertainty under Gaussian assumptions. We evaluate these metrics for a single output step, and for 12 output steps.

The Gaussian Negative Log-likelihood is used as a metric for all models, and as a loss for the baseline model. This loss only estimates aleatoric uncertainty, and is presented in Equation \ref{eq:NLL}
\begin{equation}
    \label{eq:NLL}
    \text{NLL} = N^{-1} \sum_{i=1}^N \left(\log (\sigma_i^2) + \frac{(\mu_i-y_i)^2}{\sigma_i^2}\right)
\end{equation}
Where $N$ is the number of data points, $\mu_i$ is the predicted mean, $\sigma^2_i$ is the prediction variance, and $y_i$ is the ground truth label.

Plotting metrics against the prediction horizon entails some expectations as the prediction horizon is longer (more timesteps into the future). Mean squared error, Mean absolute percentage error should increase with longer horizons, while the $R^2$ score should decrease. The calibration error should stay approximately constant, as increasing uncertainty should signal increasing error in the predictions, and NLL should stay approximately constant as this metric also considers uncertainty in its predictions.

\section{Experimental Results}

Our main numerical results are presented here. Tables \ref{tab:mlpPM25} and \ref{tab:lstmPM25} present results for the PM2.5 dataset, while tables \ref{tab:mlpPressure} and \ref{tab:lstmPressure} correspond to the Air Pressure dataset.

For PM2.5 MLP models, ensembles performed the best according to all metrics. Flipout and Baseline models were the worst fit for the data, but
the scores were rather similar, there were no outliers. Flipout and BBB performed worse in uncertainty related metrics (CE and NLL) compared to other models. For the LSTM model, results are similar as the best algorithm is still the ensemble model but DropConnect performs much better. BBB performed extremely poorly, with a negative R2 score and an MSE larger than 1. Overall for both kinds of models, BBB predicts very poor uncertainty. Among all the models, Ensemble performed the best, with a slight margin ahead of Dropout or
Dropconnect depending on the recurrent layers. Both the standard and recurrent versions of these models outperformed the other estimation methods.

\begin{table}[!t]
    \centering
    \begin{tabular}{lllllll}\hline
            Models/Metrics& MAPE $\downarrow$ & MSE $\downarrow$ & $R^2$ $\uparrow$ & Calibration error $\downarrow$ & NLL $\downarrow$ \\ \hline
            MLP Baseline & 51.85 & 0.45 & 0.56 & 0.21 & 5.57 \\ %
            MLP Ensemble & \textbf{45.29} & \textbf{0.36} & \textbf{0.65} &  \textbf{0.20} &  \textbf{1.86}\\ %
            MLP Dropout & 46.99 & 0.38 & 0.62 & 0.26 & 4.38\\ %
            MLP Dropconnect & 59.14 & 0.52 & 0.49 & 0.25 & 5.52\\ %
            MLP BBB & 47.00 & 0.37 &  0.64 & 0.37 & 29.71\\ %
            MLP Flipout & 50.14 & 0.43 & 0.57 & 0.35 & 17.84\\ \hline
    \end{tabular}
    \caption{Metrics scores of each model (PM2.5).}
    \label{tab:mlpPM25}
\end{table}

\begin{table}[!t]
    \centering
    \begin{tabular}{lllllll}
            \hline
            Models/Metrics & MAPE $\downarrow$ & MSE $\downarrow$ & $R^2$ $\uparrow$ & Calibration error $\downarrow$ & NLL $\downarrow$ \\ \hline
            LSTM Baseline 	& 50.16 & 0.46 & 0.55 & 0.28 & 8.04 \\ %
            LSTM Ensemble 	& \textbf{43.13} & \textbf{0.34} & \textbf{0.67} & \textbf{0.17} & \textbf{1.61}\\ %
            LSTM Dropout 	& 48.21 & 0.41 & 0.60 & 0.29 & 8.59\\ %
            LSTM Dropconnect & 44.78 & 0.36 & 0.65 & 0.17 & 4.33\\ %
            LSTM BBB 		& 92.48 & 1.15 & -0.11& 0.42 & 102.71\\ %
            LSTM Flipout 	& 50.14 & 0.41 & 0.59 & 0.34 & 13.85\\ \hline
    \end{tabular}
    \caption{Metrics scores of each model with recurrent layer (PM2.5).}
    \label{tab:lstmPM25}
\end{table}

For Air Pressure models, the results differ considerably from the PM2.5 dataset. For the MLP, the best model predictive model was Dropout. As before, Ensembles was the best in predicting uncertainty, since it produced the lowest NLL score and calibration error. Interestingly, the Baseline model was the best calibrated in a tie with Ensemble, which means that they were the least overconfident amongst the models. Among LSTM models, the best is again Ensembles, Dropconnect performed significantly better with LSTM. Both Flipout and BBB have really low performance both in predicting the true mean and the standard deviation. Note that BBB had really low calibration error compared to the previous cases.

In general, the models performed worse with the Air pressure than with the PM2.5 dataset. It is also worth mentioning that the Baseline model performed relatively well in all settings, especially with CE and NLL scores. Overall the best model for uncertainty was Ensemble however, in some cases, Dropout or Dropconnect outperformed it. An important detail that is often overlooked, is that our results show that a recurrent model (LSTM in this case) is not always the best model for prediction of time series data, an MLP can outperform LSTM, we see this in MLP Dropout on the Air pressure dataset. There are also interactions between quality of predicted uncertainty with LSTM, calibration error or NLL can increase or decrease, there is no general pattern.

\begin{table}[!t]
    \centering
    \begin{tabular}{lllllll}
            \hline
            Models/Metrics& MAPE $\downarrow$ & MSE $\downarrow$ & $R^2$ $\uparrow$ & Calibration error $\downarrow$ & NLL $\downarrow$ \\ \hline
            MLP Baseline & 57.19 & 0.53 & 0.36 & 0.18 & 2.62 \\ %
            MLP Ensemble & 51.37 & 0.42 & 0.49 & \textbf{0.18} &  \textbf{1.32}\\ %
            MLP Dropout & \textbf{50.18} & \textbf{0.40} & \textbf{0.51} & 0.26 & 3.07\\ %
            MLP Dropconnect & 64.17 & 0.62 & 0.25 & 0.30 & 18.84\\ %
            MLP BBB & 57.56 & 0.65 &  0.21 & 0.38 & 38.70\\ %
            MLP Flipout & 63.25 & 0.69 & 0.17 & 0.40 & 36.16\\ \hline
    \end{tabular}
    \caption{Metrics scores of each model (Air pressure).}
    \label{tab:mlpPressure}
\end{table}

\begin{table}[!t]
    \centering
    \begin{tabular}{lllllll}
            \hline
            Models/Metrics& MAPE $\downarrow$ & MSE $\downarrow$ & $R^2$ $\uparrow$ & Calibration error $\downarrow$ & NLL $\downarrow$ \\ \hline
            LSTM Baseline 	& 62.31 & 0.62 & 0.25 & 0.29 & 6.70 \\ %
            LSTM Ensemble   & \textbf{53.26} & \textbf{0.46} & \textbf{0.45} & \textbf{0.17} & \textbf{1.56}\\ %
            LSTM Dropout 	& 59.34 & 0.57 & 0.32 & 0.34 & 10.51\\ %
            LSTM Dropconnect & 54.96 & 0.48 & 0.42 & 0.25 & 4.04\\ %
            LSTM BBB 		& 101.22 & 2.18 & -1.60 & 0.23 & 196.70\\ %
            LSTM Flipout 	& 75.99 & 2.54 & -2.04 & 0.43 & 96.60\\ \hline
    \end{tabular}
    \caption{Metrics scores of each model with recurrent layer (Air pressure).
    }
    \label{tab:lstmPressure}
\end{table}

We also provide a qualitative evaluation of prediction results, presented in Tables \ref{tab:final_table_pm2.5} and \ref{tab:final_table_airPressure}. The details of how these tables are made is provided in the appendix. These tables provide a ranking of uncertainty quantification methods across both datasets and model variations. Overall we find that the baseline model, ensembles, and dropout, are the best uncertainty quantification methods overall across MLP and LSTM models.

\begin{figure}[!h]
    \centering
    \subfloat[Baseline, Ensemble, Dropout]{\includegraphics[width=0.9\textwidth]{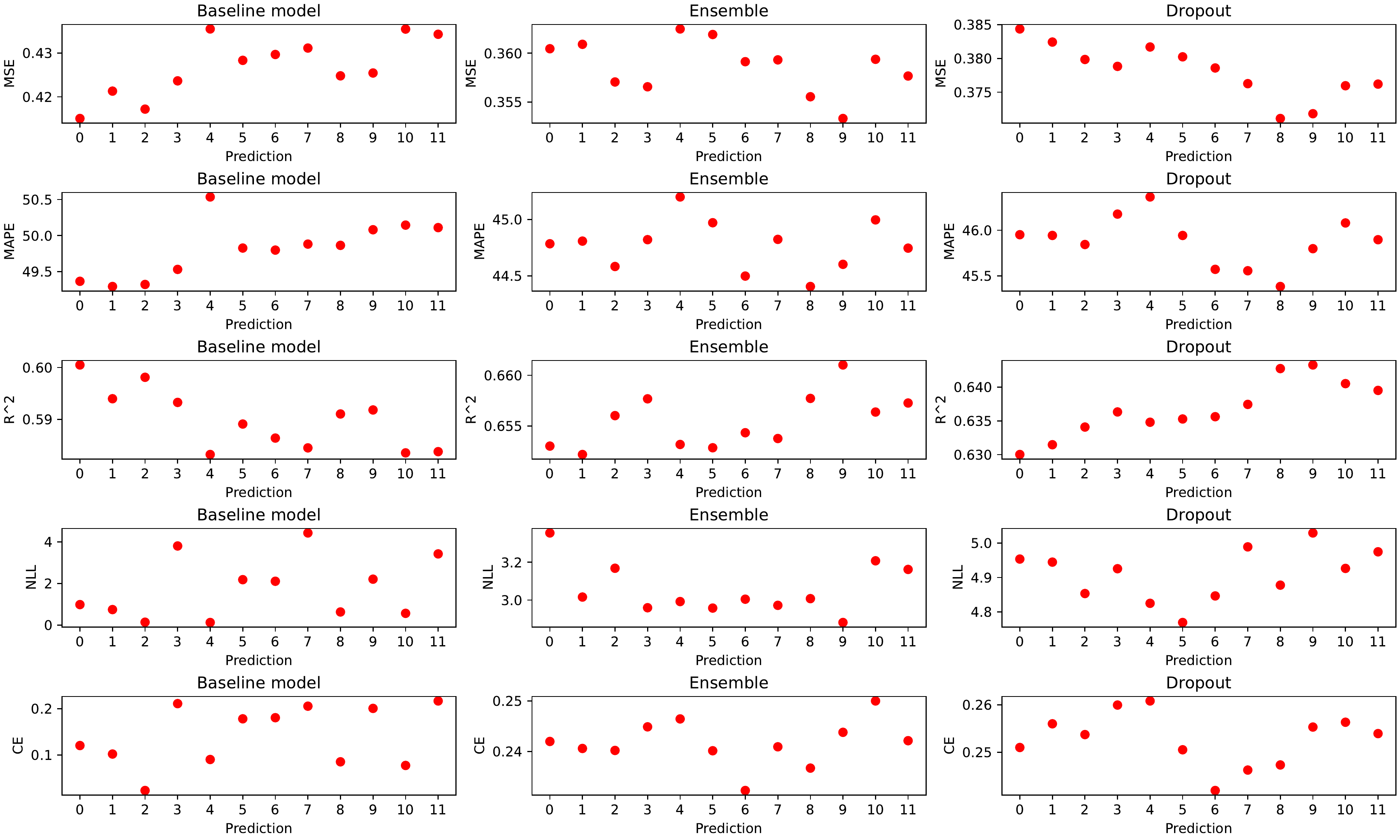}}\hfill
    \subfloat[DropConnect, BBB, Flipout.]{\includegraphics[width=0.9\textwidth]{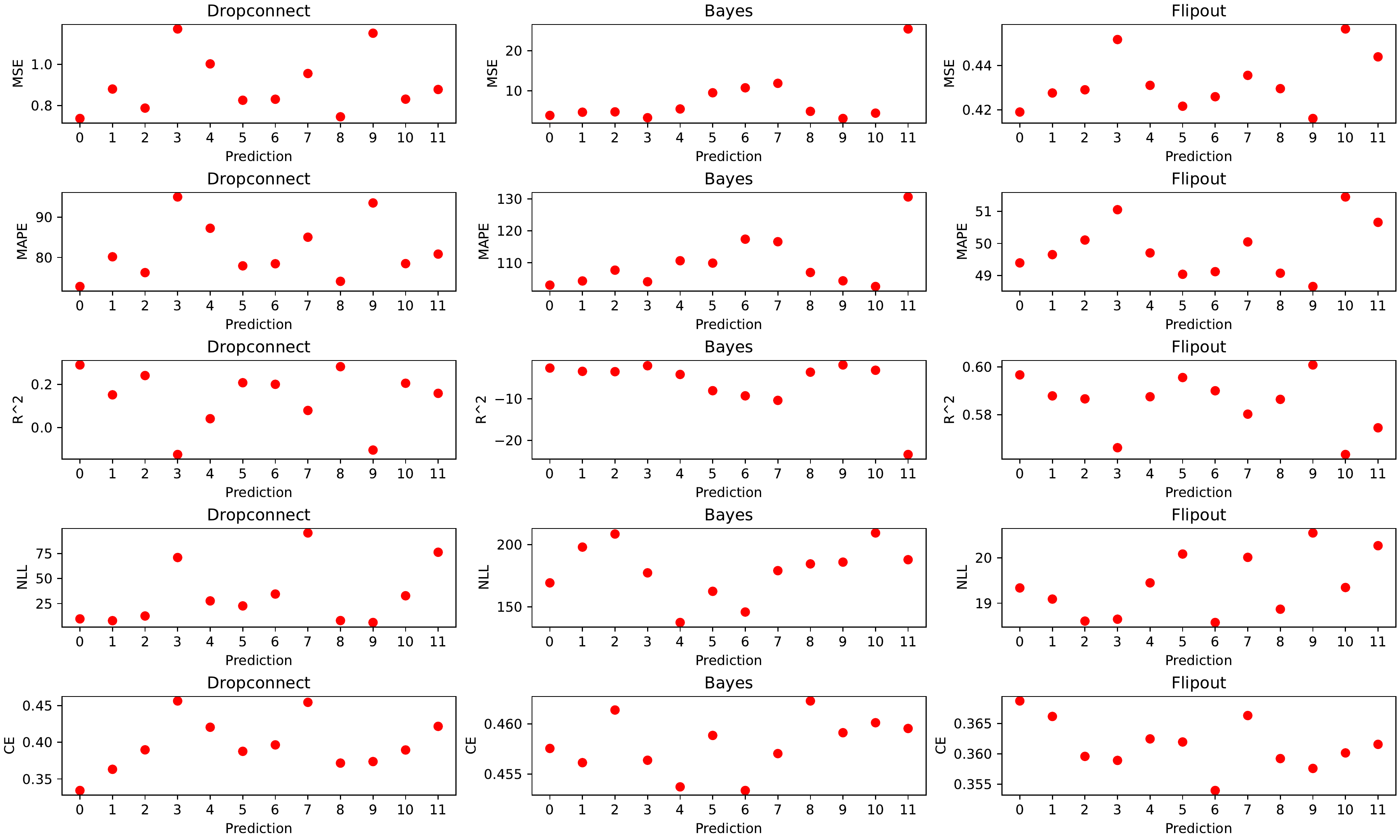}}
    \caption{Comparison of MAPE, MSE, $R^2$, CE, and NLL with different uncertainty methods for prediction horizon of the MLP model in the PM2.5 test set. The scores not always follow a gradual rise (or drop in case of  $R^2$).}
    \label{fig:HorizonMetricsDensePM25}
\end{figure}

\begin{figure}[!h]
    \centering
    \subfloat[Baseline, Ensemble, Dropout.]{\includegraphics[width=0.9\textwidth]{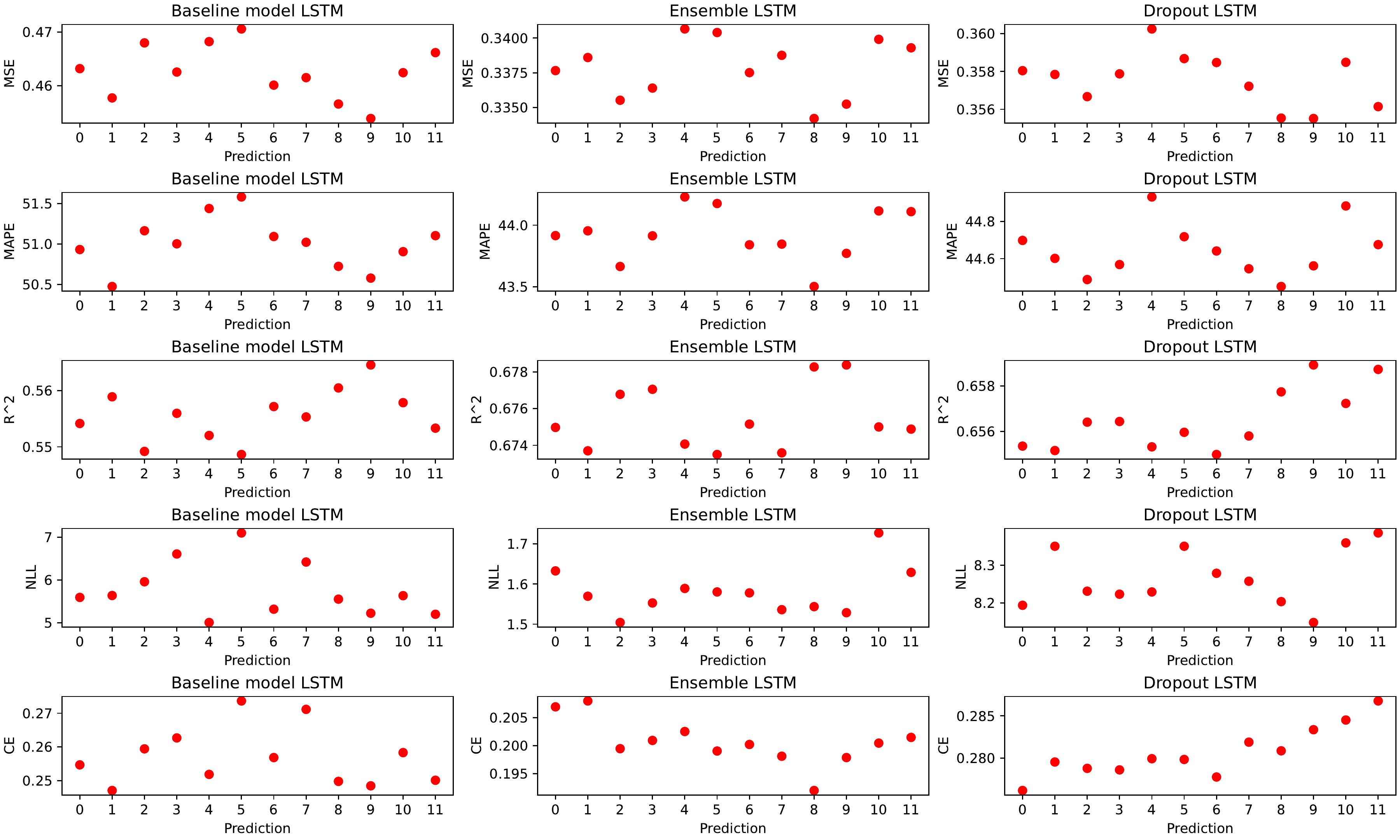}}\hfill
    \subfloat[DropConnect, BBB, Flipout.]{\includegraphics[width=0.9\textwidth]{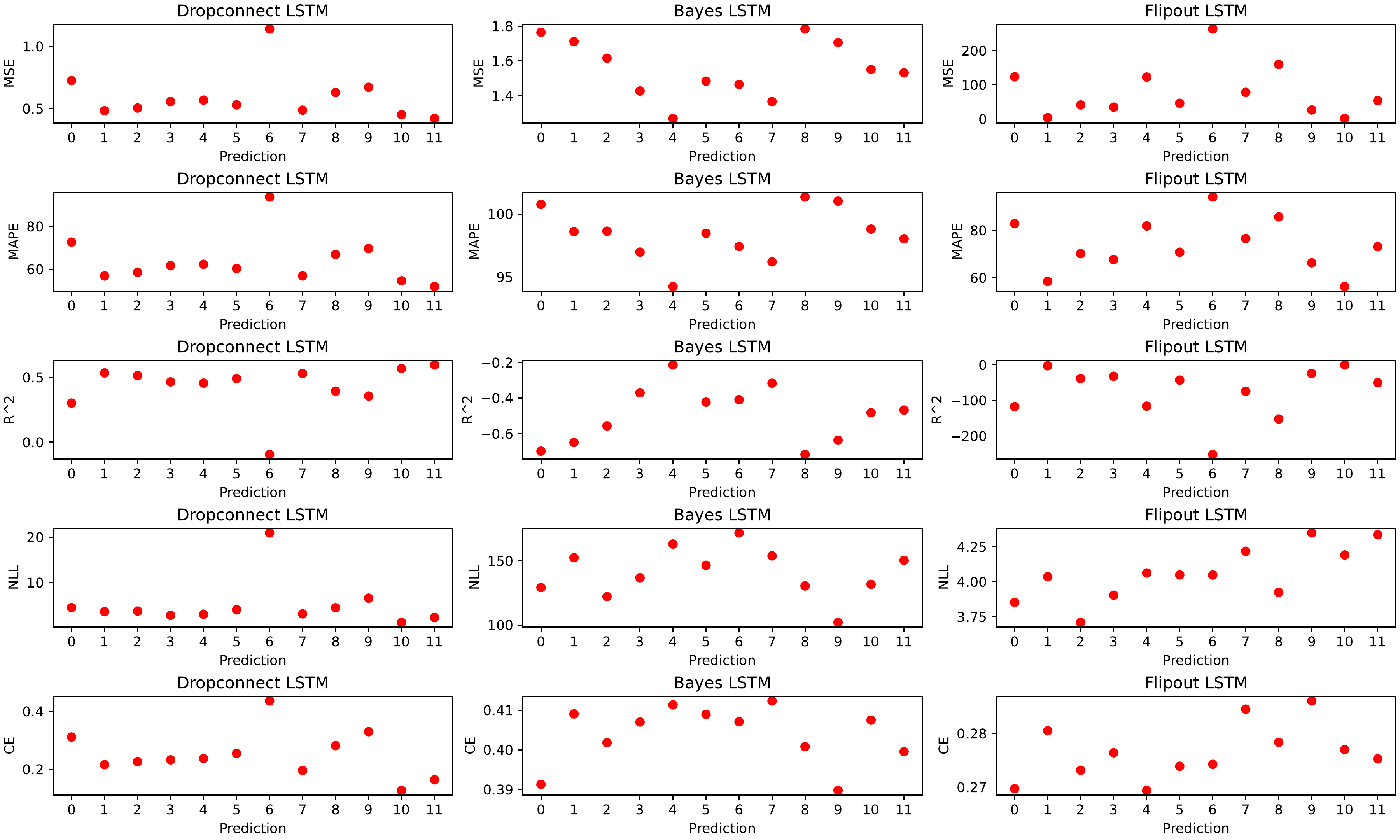}}
    \caption{Comparison of MAPE, MSE, $R^2$, CE, and NLL with different uncertainty methods for prediction horizon of the LSTM model in the PM2.5 test set. The scores not always follow a gradual rise (or drop in case of  $R^2$).}
    \label{fig:HorizonMetricsLSTMPM25}
\end{figure}

\begin{figure}[!h]
    \centering
    \subfloat[Baseline, Ensemble, Dropout]{\includegraphics[width=0.9\textwidth]{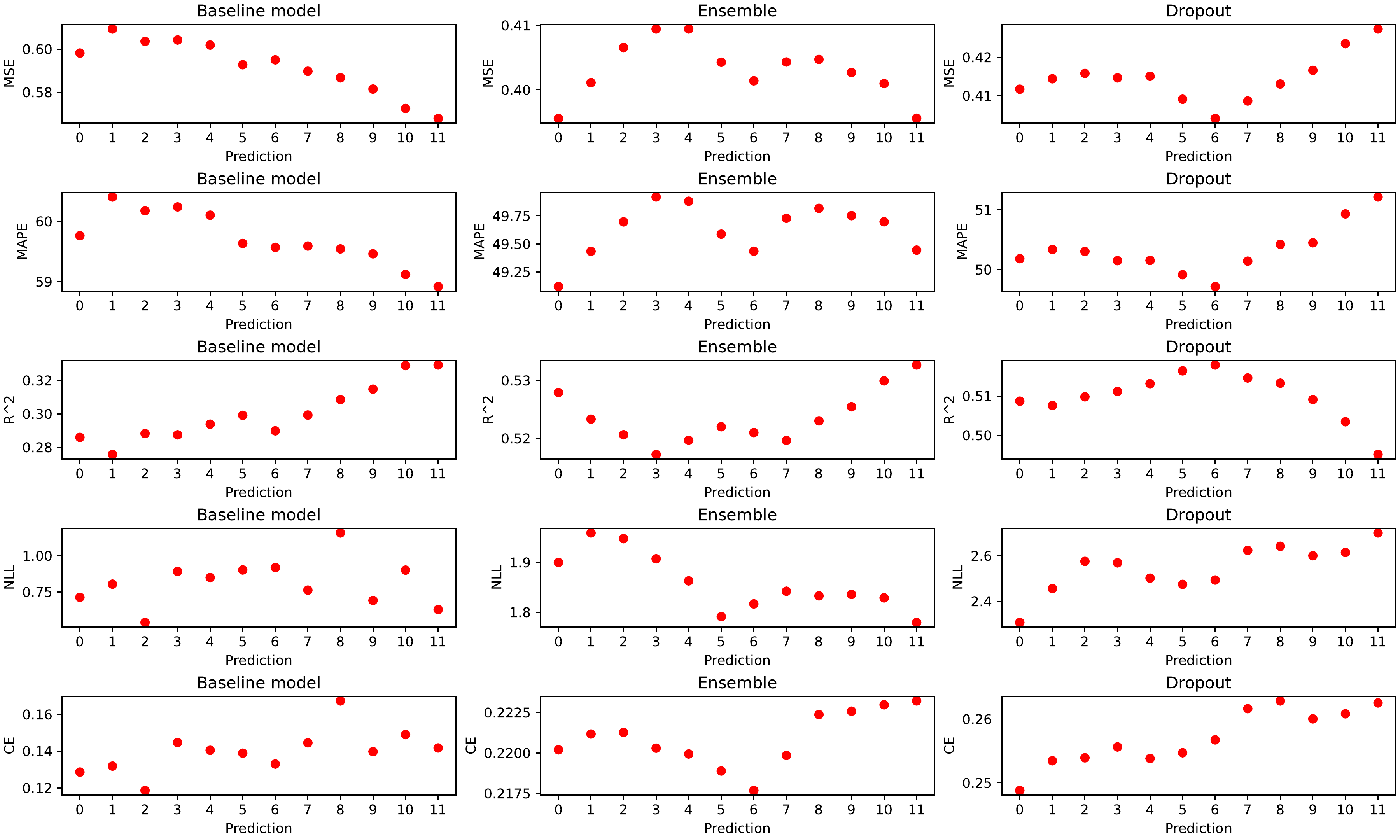}}\hfill
    \subfloat[DropConnect, BBB, Flipout.]{\includegraphics[width=0.9\textwidth]{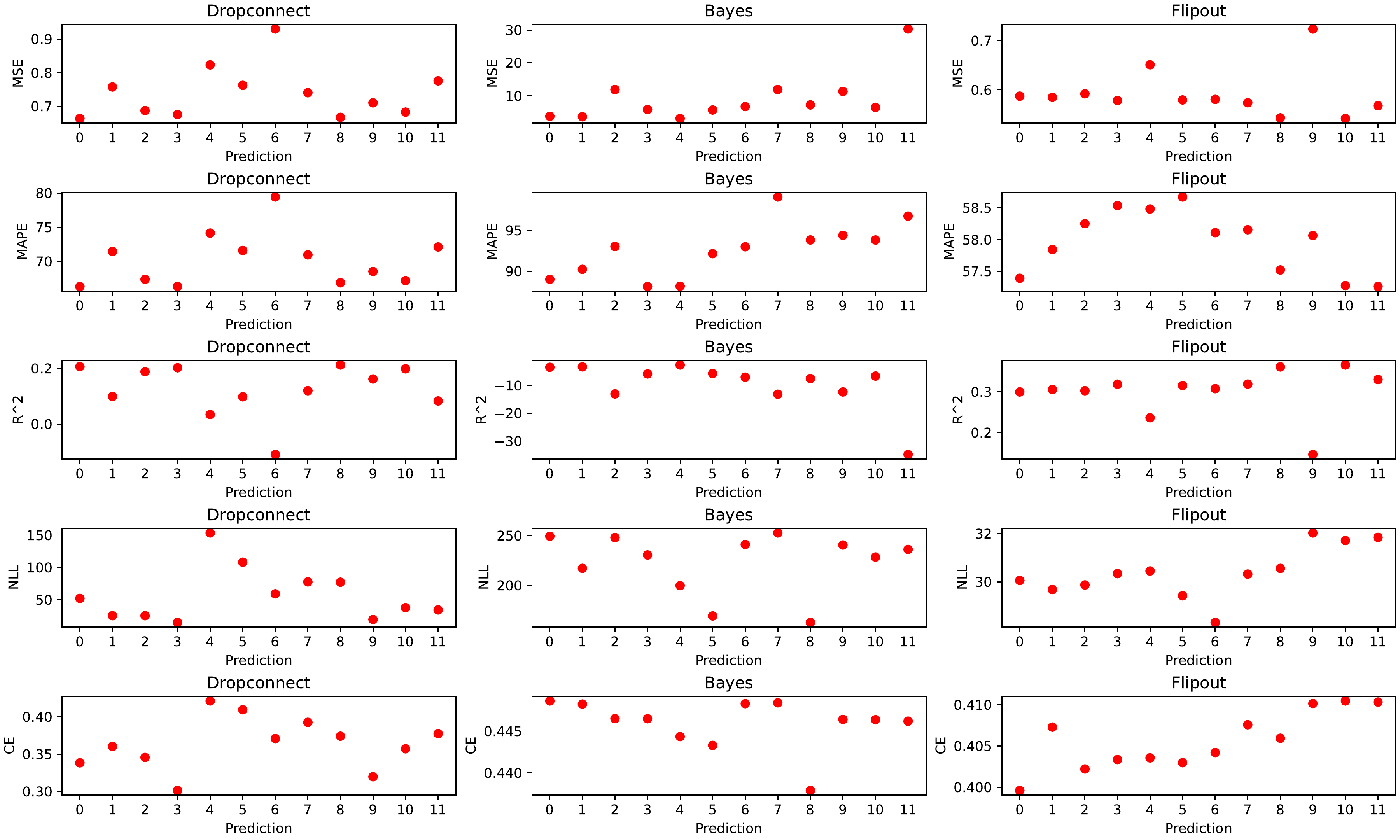}}
    \caption{Comparison of MAPE, MSE, $R^2$, CE, and NLL with different uncertainty methods for prediction horizon of the MLP model in the Air test set. The scores not always follow a gradual rise (or drop in case of  $R^2$).}
    \label{fig:HorizonMetricsDenseAir}
\end{figure}

\begin{figure}[!h]
    \centering
    \subfloat[Baseline, Ensemble, Dropout.]{\includegraphics[width=0.9\textwidth]{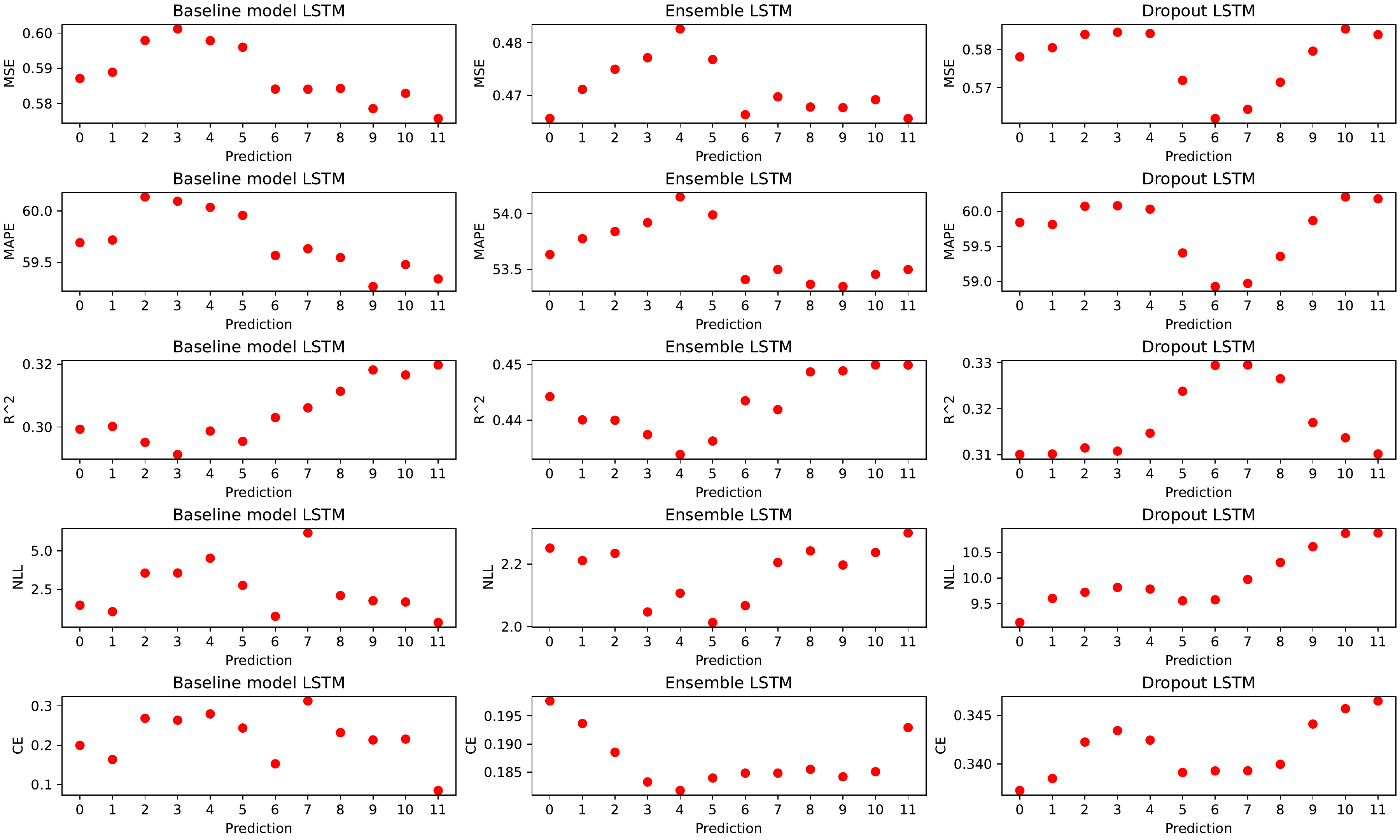}}\hfill
    \subfloat[DropConnect, BBB, Flipout.]{\includegraphics[width=0.9\textwidth]{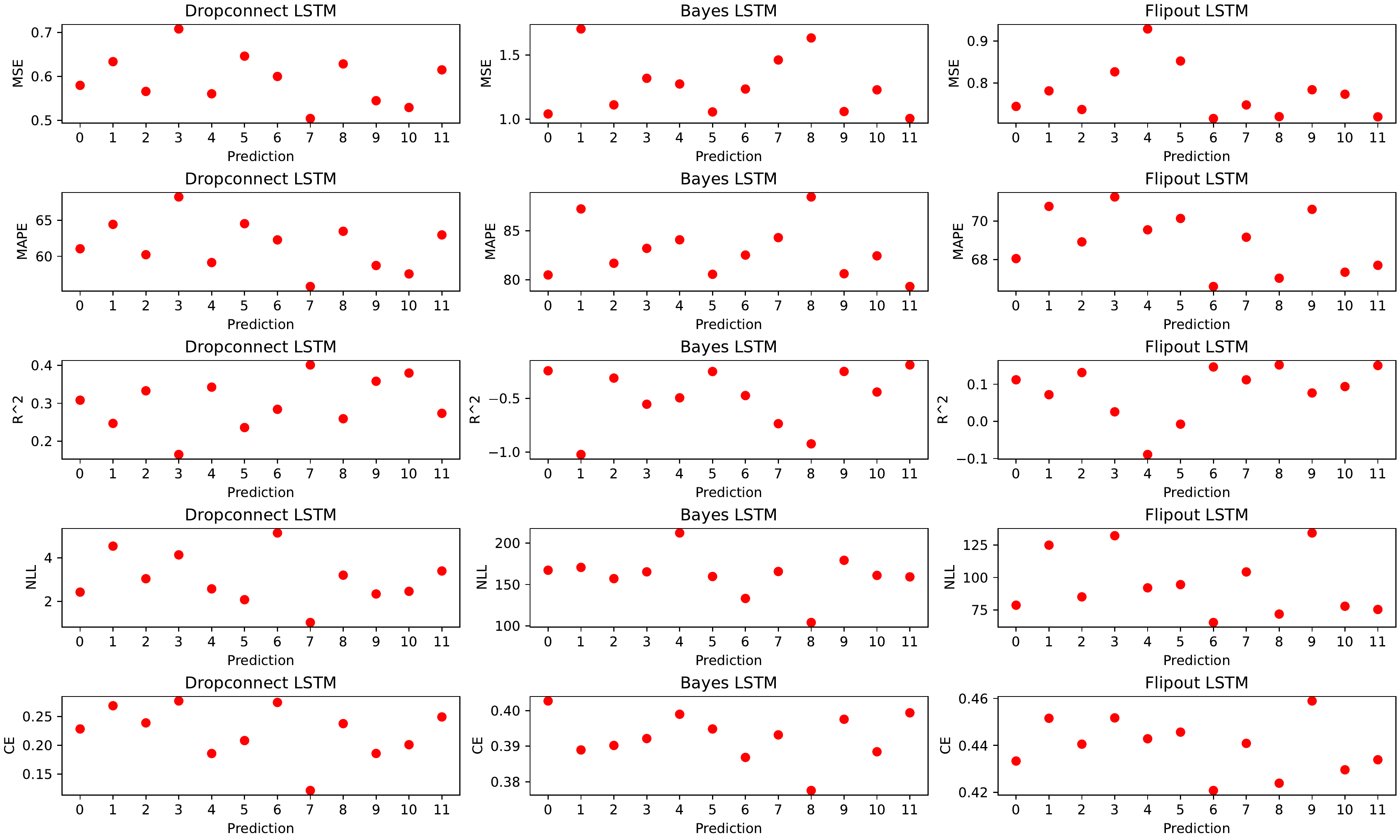}}
    \caption{Comparison of MAPE, MSE, $R^2$, CE, and NLL with different uncertainty methods for prediction horizon of the LSTM model in the Air test set. The scores not always follow a gradual rise (or drop in case of  $R^2$).}
    \label{fig:HorizonMetricsLSTMAir}
\end{figure}

Metrics across prediction horizon are presented in Figures \ref{fig:HorizonMetricsDensePM25} to \ref{fig:HorizonMetricsLSTMAir}. These results are also analyzed in the qualitative results (Sec \ref{sec:modelRanking}). Overall the results are mixed, some model/uncertainty method combinations do meet the expectations, such as PM2.5 MLP Dropout, and Air Pressure LSTM Dropout. Overall our results in this category show that performing time series regression with uncertainty is far from trivial and many assumptions usual in sequence data can be broken.

Qualitative evaluation of the prediction horizon quality and confidence vs error plots shows that the baseline models, ensembles, and droput are the best overall, but still they do violate some assumption in some cases, no model fills all expectations in all metrics.

\begin{table}[!h]
    \centering
    \begin{tabular}{llllllll}
        \toprule
        Data & \multicolumn{7}{c}{PM2.5}\\
        \midrule
        Models/Metrics&  MAPE & MSE  & $R^2$  & Calib error  & NLL & Horizon & Conf vs Error\\
        \midrule
        \midrule
        MLP Baseline & 10 & 9 & 9 & 4 & 6 & \textbf{Good} & Bad\\ %
        \textbf{MLP Ensemble} &  \textbf{3} & \textbf{3} & \textbf{3} & \textbf{3} & \textbf{2}&  \textbf{Good} & Moderate \\
        MLP Dropout &  4 & 5 & 5 & 6 & 4 & Moderate & Bad \\ %
        MLP Dropconnect &  11 & 11 & 11 & 5 & 5 & Bad & Moderate\\ %
        MLP BBB & 5 & 4 & 4 & 11 & 11 & Moderate & Bad  \\ %
        MLP Flipout &  7 & 8 & 8 & 10 & 10 & Bad & Bad \\ %
        \midrule
        LSTM Baseline  & 9 & 10 & 10 & 7 & 7 & Bad & \textbf{Good}\\
        \textbf{LSTM Ensemble}&  \textbf{1} & \textbf{1} & \textbf{1} & \textbf{1} & \textbf{1}& Bad & \textbf{Good}\\ %
        LSTM Dropout &  6 & 6 & 6 & 8 & 8 & Bad & Moderate\\
        \textbf{LSTM Dropconnect}&  \textbf{2} & \textbf{2} & \textbf{2} & \textbf{2} & \textbf{3}& Bad & Moderate\\ %
        LSTM BBB  &  12 & 12 & 12 & 12 & 12 & Bad & Moderate\\
        LSTM Flipout  &  8 & 7 & 7 & 9 & 9 & Bad & Moderate\\
        \bottomrule
    \end{tabular}
    \caption{Ranking of models regarding the different metrics/tasks for the PM2.5 data. The Ensemble models and the LSTM version of the Dropconnect model performed the best. For the horizon check, only the Baseline model performed in the desired manner. Regarding the Confidence vs Error plots, the LSTM Baseline and Ensemble models performed as expected.}
    \label{tab:final_table_pm2.5}%
\end{table}

\begin{table}[!h]
    \vspace{7mm}
    \centering
    \begin{tabular}{llllllll}
        \toprule
        Data & \multicolumn{7}{c}{Air pressure}\\
        \midrule
        Models/Metrics&   MAPE & MSE  & $R^2$  & Calib error  & NLL & Horizon& Conf vs Error \\
        \midrule
        \midrule
        MLP Baseline & 5 & 5 & 5 & \textbf{3} & \textbf{3} & Bad & Moderate\\ %
        \textbf{MLP Ensemble} &  \textbf{2} & \textbf{2} & \textbf{2} & \textbf{2} & \textbf{1}& Moderate  & Bad\\ %
        \textbf{MLP Dropout} &  \textbf{1} & \textbf{1} & \textbf{1} & 6 & 4& \textbf{Good} & Moderate\\ %
        MLP Dropconnect &  10 & 8 & 8 & 8 & 8 & Bad & Bad\\ %
        MLP BBB & 6 & 9 & 9 & 10 & 10& Moderate & Moderate  \\ %
        MLP Flipout &  9 & 10 & 10 & 11 & 9& Moderate & Bad\\ %
        \midrule
        LSTM Baseline & 8 & 7 & 7 & 7 & 6 & Bad & \textbf{Good}\\ %
        \textbf{LSTM Ensemble}&  \textbf{3} & \textbf{3} & \textbf{3} & \textbf{1} & \textbf{2}& Bad & \textbf{Good}\\ %
        LSTM Dropout &  7 & 6 & 6 & 9 & 7& \textbf{Good} & \textbf{Good}\\ %
        LSTM Dropconnect &  4 & 4 & 4 & 5 & 5& Bad & Bad\\ %
        LSTM BBB &  12 & 11 & 11 & 4 & 12& Moderate & Moderate\\ %
        LSTM Flipout &  11 & 12 & 12 & 12 & 11& Bad & Moderate\\
        \bottomrule
    \end{tabular}
    \caption{Ranking of models regarding the different metrics/tasks for the Air pressure data. The Ensemble models and the Dropout model performed the best in general. For the uncertainty based metrics, the Baseline model was successful as well. Regrading the horizon check, the Dropout models significantly outperformed all the other model. Regarding the Confidence vs Error plots, the LSTM Baseline, Ensemble and Dropout models performed as expected.}
    \label{tab:final_table_airPressure}%
\end{table}

\section{Conclusions and Future Work}

In this work we have made a comprehensive evaluation of uncertainty quantification methods for time series data, on two regression datasets (PM2.5 and Air Pressure prediction), considering multiple factors like prediction quality, uncertainty quality, and in particular how the prediction horizon relates with predicted uncertainty. It was expected that uncertainty and error increase with predictions far into the future, but overall this does not always hold. It strongly depends on the uncertainty method and the base model (recurrent or feed forward).

The expectations of uncertainty when it comes to time series data can be partially fulfilled, but the choice of model is crucial. Ensemble and Dropout/Dropconnect models performed the best. However, the model selection is highly dependent on the dataset and task. The dataset and the model also influence the need for recurrent layers. For example, they outperformed the standard models in Confidence vs Error expectation but the horizon expectation worked better for the non-recurrent models. 

For future work, we believe that developing uncertainty quantification methods that are directly designed for sequence and time series data is crucial. All the current methods were not directly designed for sequence data, they can be partially adapted, but overall these methods do not have built-in intuitions for long and short term in-sequence dependencies, which might improve performance.

\clearpage
\FloatBarrier
\bibliographystyle{plain}
\bibliography{literature}

\FloatBarrier
\clearpage
\appendix
\section{Overall Ranking of Models}
\label{sec:modelRanking}

In this section we provide a qualitative evaluation of each uncertainty quantification method across the two kinds of models (MLP and LSTM) and the two datasets. These results were presented in Tables \ref{tab:final_table_pm2.5} and \ref{tab:final_table_airPressure}. In this section we define a set of expectations/criteria for how uncertainty should behave when it comes to prediction for time series data:
\begin{enumerate}
    \item A wider prediction horizon should result in higher uncertainty.
    \item Larger errors (for instance, mean squared error) should yield higher uncertainty.
    \item The estimation of calibration error (whether the model is over or under-confident). This suggests that the standard deviation (in other words, the uncertainty) should be proportional to the accuracy of the model.
\end{enumerate}
Then we define two categorical parameters. The first is about the prediction horizon quality:

\begin{tabular}{ll}
    \toprule
    \textbf{Bad} 		& where there were no conclusive results. \\
    \textbf{Moderate} 	& when there were at least two metrics from which satisfied the expectations.\\
    \textbf{Good}		& if three expectations are satisfied.\\
    \bottomrule
\end{tabular}

And the second about Confidence vs Error plots expectation, in this case:

\begin{tabular}{ll}
    \toprule
    \textbf{Bad} 		& the error does not increase with larger uncertainty.\\
    \textbf{Moderate} 	& the error increases with uncertainty but it fluctuates at higher uncertainty.\\
    \textbf{Good}		& the error increases with uncertainty and does not oscillate.\\
    \bottomrule
\end{tabular}
\FloatBarrier

\subsection{Uncertainty Quantification Methods}

This section provides a short description of the uncertainty quantification methods we evaluated, including the hyper-parameter values we used for training.

\textbf{MC-Dropout}. Dropout randomly sets layer activations to zero with probability $p$, and it was originally intended as a regularizer that is only applied during training. MC-Dropout \cite{gal} enables the dropping of activations at during test/inference time, making the model stochastic. It has been shown that each forward pass produces one sample from the corresponding Bayesian posterior distribution \cite{gal}. We use drop probability $p = 0.2$.

\textbf{MC-DropConnect}. DropConnect is similar to Dropout, randomly dropping weights to zero with probability $p$ instead of activations. MC-DropConnect enables this behavior at inference time, which also produces samples from the Bayesian posterior distribution \cite{mobiny2021dropconnect}. We use a drop probability $p = 0.05$, note that this is smaller than the drop probability as it is applied to weights, using a larger probability reduces the model capacity proportionally.

\textbf{Ensembles}. Consist of training multiple copies of the same architecture and then combining their predictions, which usually produces a better model. Lakshminarayanan \cite{lakshminarayanan2017simple} demonstrated that ensembles also have good uncertainty quantification properties. We use an ensemble of $M = 10$ neural networks of same architecture.

\textbf{Bayes by Backprop (BBB)}. This is a full Bayesian neural network, where each weight is approximated as a Gaussian distribution $\condprob{\mathbf{w}}{\mathbf{x}}$ using variational inference \cite{Charles}. The model is stochastic, each forward pass produces on sample, and the predictive posterior distribution $\condprob{\mathbf{y}}{\mathbf{x}}$ (Eq \ref{ppd}) is approximated through Monte Carlo sampling (Eq \ref{mc-ppd}) with $M = 50$ forward passes.

\begin{equation}
    \condprob{\mathbf{y}}{\mathbf{x}} = \int_\mathbf{w}\condprob{\mathbf{y}}{\mathbf{w}, \, \mathbf{x}} \condprob{\mathbf{w}}{\mathbf{x}} \, d\mathbf{w}
    \label{ppd}
\end{equation}    

\begin{equation}
    \condprob{\mathbf{y}}{\mathbf{x}} \sim M^{-1} \sum_i^M \mathbb{P}_i(\mathbf{y} \, | \, \mathbf{w}, \mathbf{x})
    \label{mc-ppd}
\end{equation}

\textbf{Flipout}. Flipout is a variation of Bayes by Backprop that is used to reduce the training process variance, improving learning stability and performance. This is done by sampling the kernel and bias matrices for each sample in a batch, through some mathematical tricks, while BBB samples one kernel and bias matrix for the whole batch, slowing down learning performance and convergence speed. Overall Flipout performs better than BBB. We use the same $M = 50$ number of forward passes as BBB and output distributions are estimated in the same way.

\section{Broader Impact Statement}

Neural networks can have a large impact on society, specially as models are increasingly used for real-world applications involving humans. In this paper about the combination of sequence data/time series and uncertainty quantification, we note the following misconceptions or possibly negative impacts:

\begin{itemize}
    \item Uncertainty or standard deviation of model output should be used to decide if the prediction should be trusted, if uncertainty is too high, predictions can be rejected, but this has to be implemented in the system using the machine learning model. This is one of the major advantages of uncertainty, but it must be used to derive highly uncertain or ambiguous predictions to a human decision maker.
    \item Uncertainty values produced by machine learning models are only approximations of the true bayesian posterior distribution, thus these standard deviation predictions should not be blindly trusted, there are no strict guarantees that uncertainty is proportional to error, as results in our paper also show. These values should be used with care, specially in out of distribution settings.
    \item Time series data might contain structure across sequence/timesteps that is not obvious and might inadvertently leak user information, for example, revealing behavior changes over time, or leaking sensitive information that can be accumulated over time. Our paper does not study how output uncertainty could contribute to this effect.
\end{itemize}

\section{Error vs Confidence Plots}

In this section we present all reliability plots and error vs confidence plots for both datasets. The error vs confidence plot is used to evaluate the quality of uncertainty in Section \ref{sec:modelRanking}.

To produce a error vs confidence plot, we take standard deviation values over a dataset $\sigma$, and iterate over a discretization $\sigma_t \in [\min \sigma, \max \sigma]$ with a predefined number of steps $S$, computing the error of predictions where $\sigma > \sigma_t$, and then plot the prediction error versus the min-max normalized value of $\sigma_t$. In our plots, the mean absolute error is used to have a comparable scale with the standard deviation output of the model.

The idea of the error vs confidence plot is to compare the prediction error as it changes with its predicted uncertainty (standard deviation), as predictions with increasing standard deviation should correspondingly have a larger error, while predictions with low standard deviation should also have low error. This can be evaluated with the calibration plot, but the error vs confidence plot gives an alternative perspective due to different grouping of predictions by uncertainty.

\begin{figure}[!hb]
    \centering
    \subfloat[MLP models]{\includegraphics[width=0.7\textwidth]{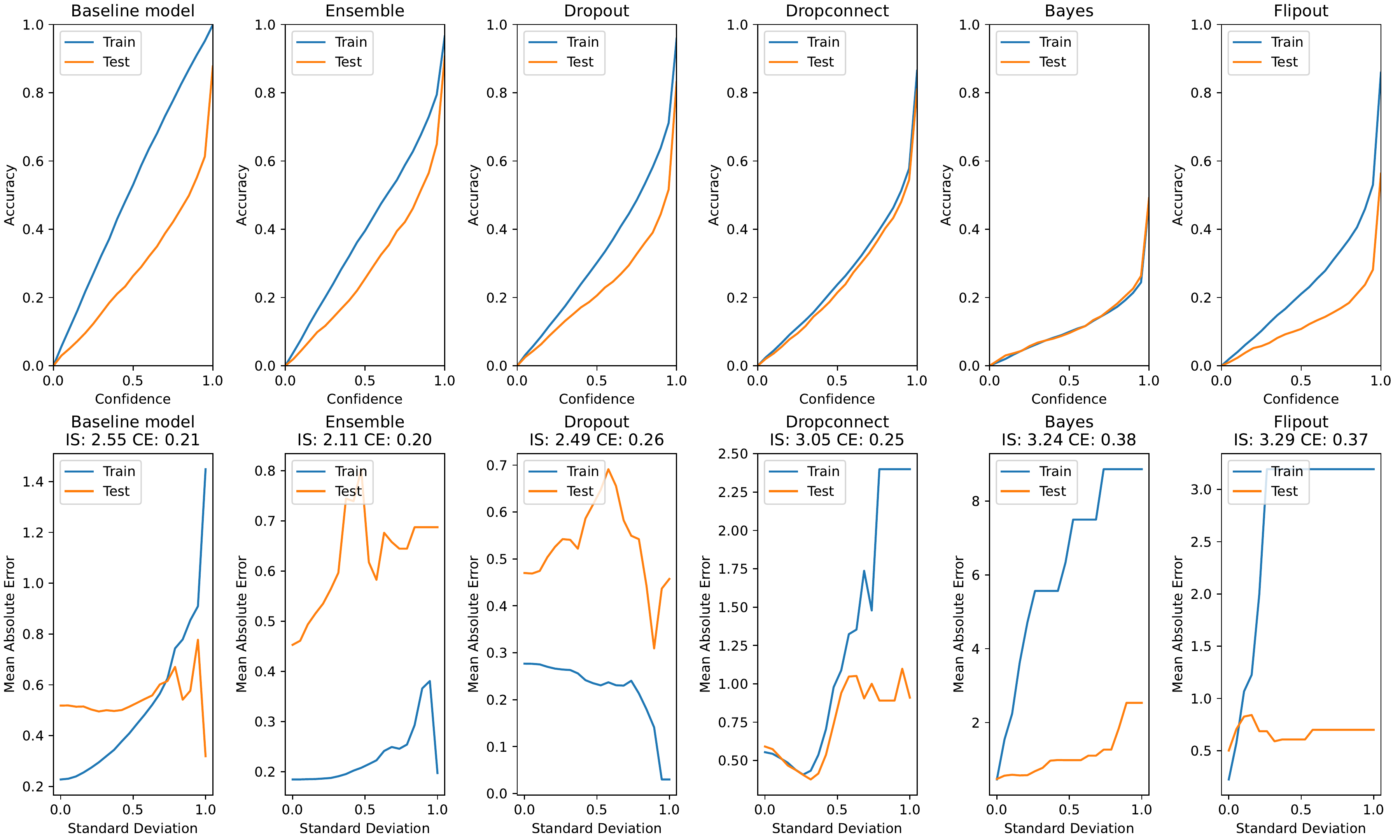}}\hfill
    \subfloat[LSTM models]{\includegraphics[width=0.7\textwidth]{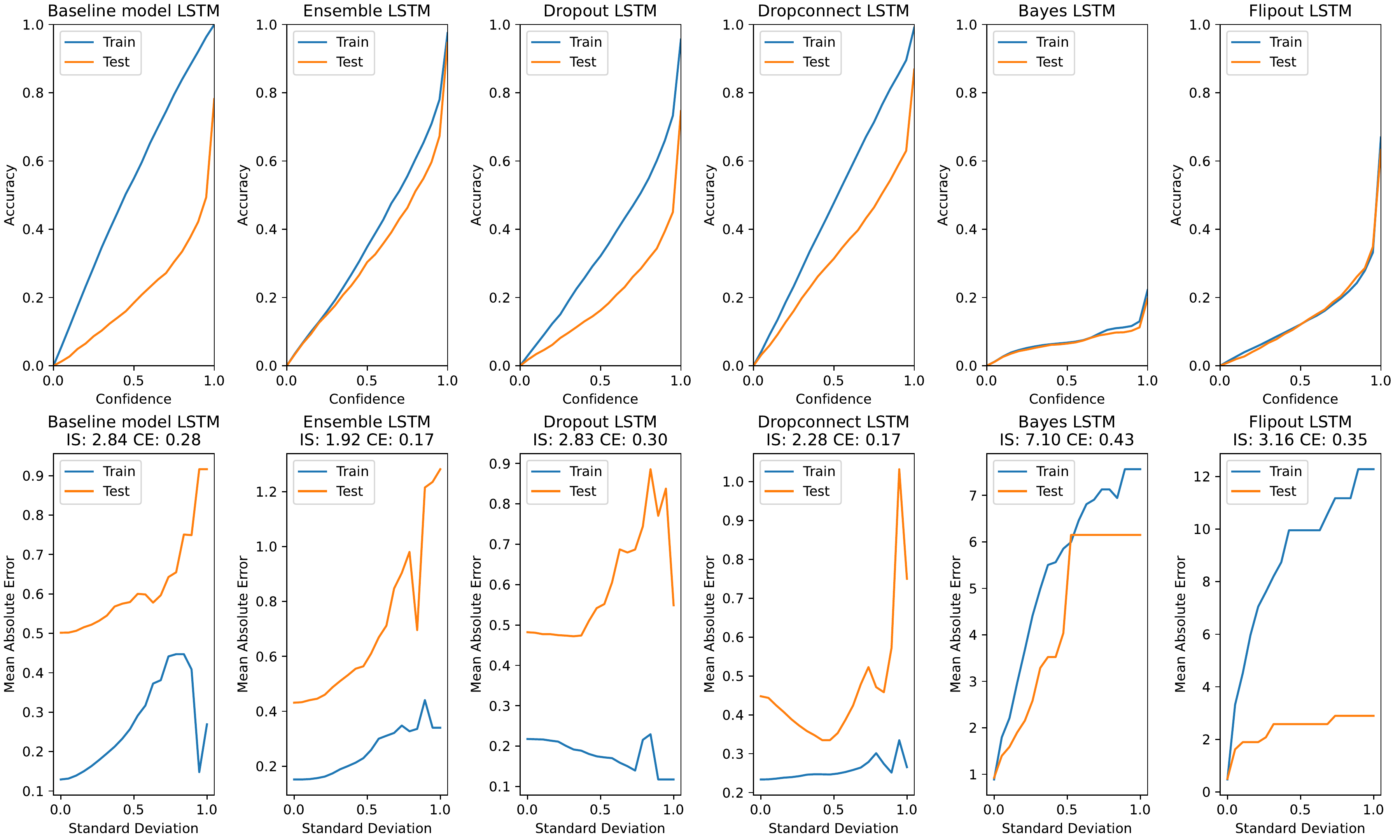}}
    \caption{For MLP and LSTM models, we present the calibration/reliability plots (top row) and confidence vs error plots (bottom row) on the PM2.5 dataset.}
    \label{fig:calibConfErrPM25}
\end{figure}

\begin{figure}[!hb]
    \centering
    \subfloat[MLP models]{\includegraphics[width=0.7\textwidth]{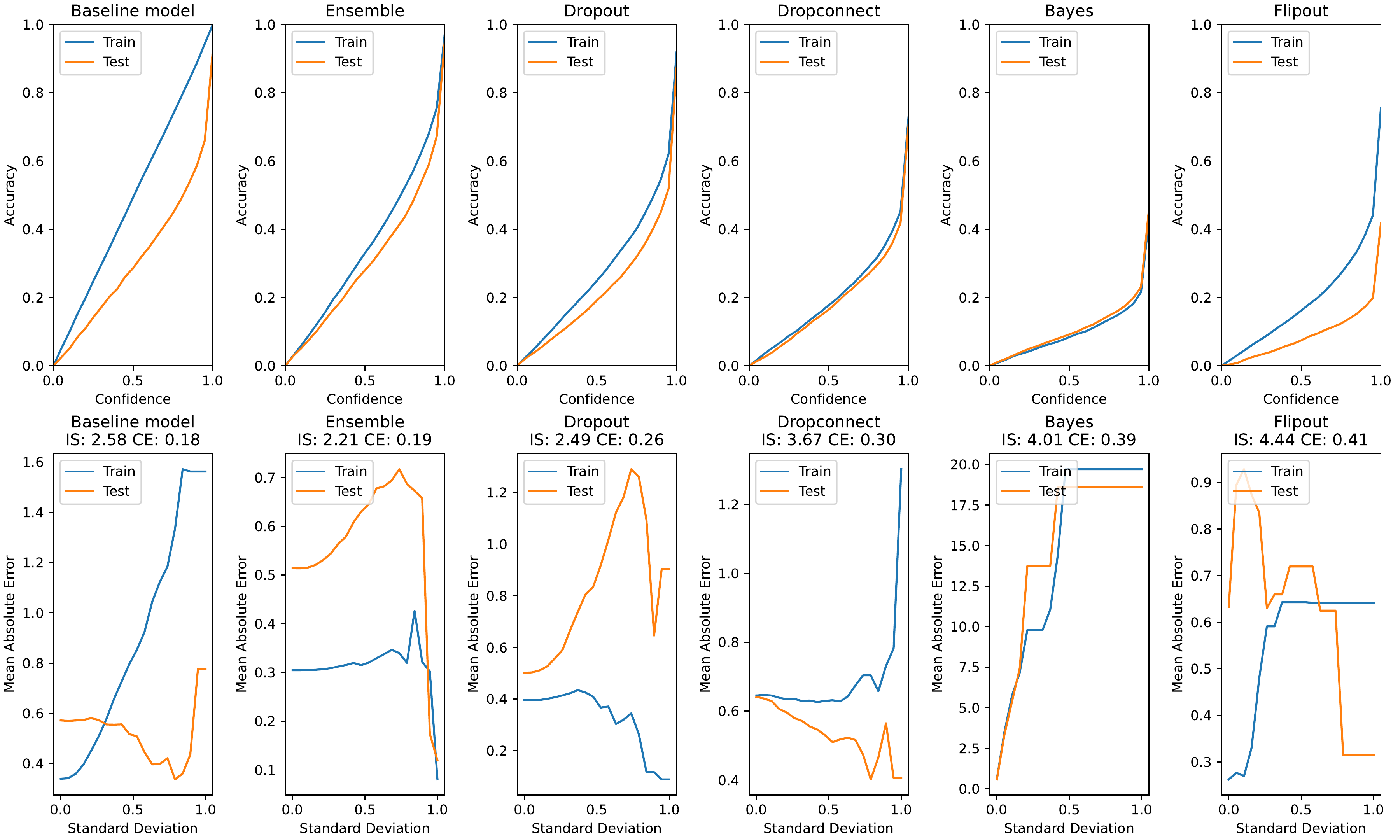}}\hfill
    \subfloat[LSTM models]{\includegraphics[width=0.7\textwidth]{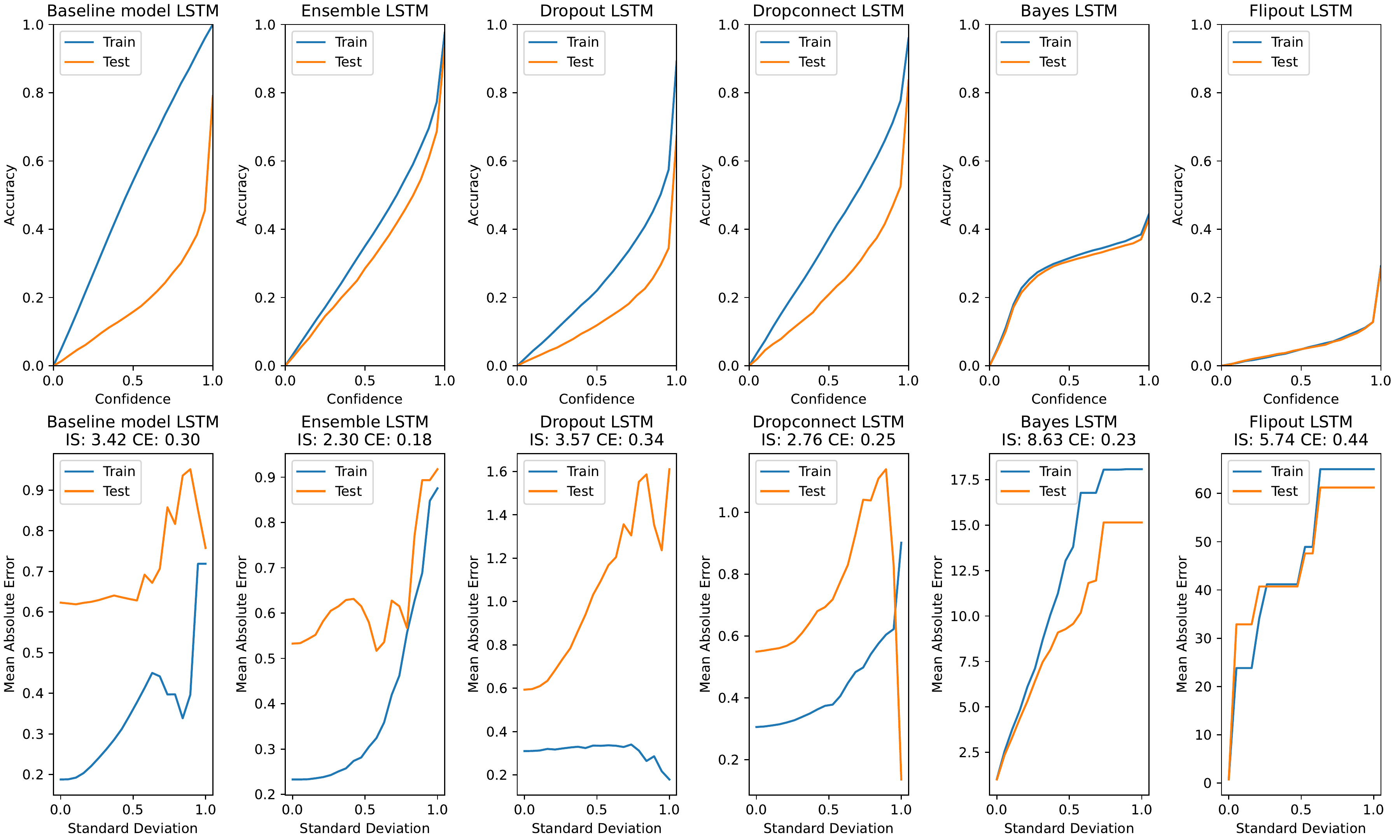}}
    \caption{For MLP and LSTM models, we present the calibration/reliability plots (top row) and confidence vs error plots (bottom row) on the Air pressure dataset.}
    \label{fig:calibConfErrAirPressure}
\end{figure}

\clearpage
\FloatBarrier

\section{Visualization of Wide Horizon Results}

This section presents a visualization of all metrics with a variable prediction horizon, up to 12 timesteps into the future. These results are used in Section \ref{sec:modelRanking} to rank models according to quality of their prediction horizon, according to the expectation that uncertainty (standard deviation) should increase with longer prediction horizons (it is harder to predict the future, specially the far future), and this should also be reflected in error and other relevant metrics.

\subsection{PM2.5 Dataset}

\FloatBarrier
\subsection{Air Dataset}

\end{document}